\documentclass[%
 aip,
 amsmath,amssymb,
 reprint,%
]{revtex4-1}

\usepackage{graphicx}
\usepackage{dcolumn}
\usepackage{bm}

\usepackage[utf8]{inputenc}
\usepackage[T1]{fontenc}
\usepackage{mathptmx}
\usepackage{multirow}
\begin{document}

\preprint{AIP/123-QED}

\title{CLPVG: Circular limited penetrable visibility graph as a new network model for time series}

\author{Qi Xuan}
\affiliation{%
Institute of Cyberspace Security, Zhejiang University of Technology, Hangzhou 310023, China.
}%
\affiliation{%
College of Information Engineering, Zhejiang University of Technology, Hangzhou 310023, China.
}%
 \author{Jinchao Zhou}%
\affiliation{%
Institute of Cyberspace Security, Zhejiang University of Technology, Hangzhou 310023, China.
}%
\affiliation{%
College of Information Engineering, Zhejiang University of Technology, Hangzhou 310023, China.
}%
\author{Kunfeng Qiu}
\affiliation{%
Institute of Cyberspace Security, Zhejiang University of Technology, Hangzhou 310023, China.
}%
\affiliation{%
College of Information Engineering, Zhejiang University of Technology, Hangzhou 310023, China.
}%
 \author{Dongwei Xu}
 \email{dongweixu@zjut.edu.cn.}
 \affiliation{%
Institute of Cyberspace Security, Zhejiang University of Technology, Hangzhou 310023, China.
}%
\affiliation{%
College of Information Engineering, Zhejiang University of Technology, Hangzhou 310023, China.
}%
 \author{Shilian Zheng}
 \affiliation{%
Science and Technology on Communication Information Security Control Laboratory, Jiaxing 314033, China.
}%

 \author{Xiaoniu Yang}
\affiliation{%
Institute of Cyberspace Security, Zhejiang University of Technology, Hangzhou 310023, China.
}%
 \affiliation{%
Science and Technology on Communication Information Security Control Laboratory, Jiaxing 314033, China.
}%

\date{\today}

\begin{abstract}
Visibility Graph (VG) transforms time series into graphs, facilitating signal processing by advanced graph data mining algorithms. In this paper, based on the classic Limited Penetrable Visibility Graph (LPVG) method, we propose a novel nonlinear mapping method named Circular Limited Penetrable Visibility Graph (CLPVG). The testing on degree distribution and clustering coefficient on the generated graphs of typical time series validates that our CLPVG is able to effectively capture the important features of time series and has better anti-noise ability than traditional LPVG. The experiments on real-world time-series datasets of radio signal and electroencephalogram (EEG) also suggest that the structural features provided by CLPVG, rather than LPVG, are more useful for time-series classification, leading to higher accuracy. And this classification performance can be further enhanced through structural feature expansion by adopting Subgraph Networks (SGN). All of these results validate the effectiveness of our CLPVG model.
\end{abstract}

\maketitle

\textbf{Time series and graphs are two popular ways to represent big data in various areas, including sociology, biology, and technology. A lot of algorithms are being developed to process these kinds of data more effectively and efficiently. Quite recently, Visibility Graph (VG) is proposed to map time series into graphs, so as to facilitate the understanding and analysis of time series by complex network theory and graph data mining. Since then, a series of new methods, such as Limted Penetrable Visibility Graph (LPVG), are proposed to construct graphs from time series. In this paper, based on LPVG, we introduce circle system for the first time in the construction of VG and propose Circular Limited Penetrable Visibility Graph (CLPVG). Comprehensive experiments on both artificial and real-world datasets validate the effectiveness of CLPVG in resisting noises, achieving better performance of time-series classification than LPVG, especially when Subgraph Network (SGN) is adopted to expand the structural feature space. We hope our work can trigger a burst of study on visibility graph, so as to enhance our understanding of complex time series and further develop advanced graph data mining algorithms for signal processing.}

\section{\label{introduction}introduction}
Time series are very popular in real world, extracting the hidden information of which is the key to understand many complex systems in various areas including Internet, communication, biology, finance, etc. 

In recent years, a variety of machine learning methods have been applied in time series classification. Traditionally, time series are taken directly as the input to train a machine learning model~\cite{karim2017lstm}. Due to the quick development of computer vision and network science, time series are also mapped into images or graphs, so that deep learning algorithms, such as Convolution Neural Network (CNN)~\cite{wang2015imaging} and Graph Neural Network (GNN)~\cite{scarselli2008graph} can be adopted for time series classification. In this paper, we focus on mapping time series to graphs.


Visibility Graph (VG)~\cite{lacasa2008time} is considered as the first model to mapping time series to graphs, which used to be a classic method in computational geometry and robot motion. The proposal of VG model opened a new door for time series analysis, with the structure of the constructed graphs inheriting some important properties of time series. After that, a series of mapping rules were proposed to establish various networks for time series. Fore instance, Lacasa et al.~\cite{luque2009horizontal} proposed a geometrically simpler algorithm, the Horizontal Visibility Graphs (HVG), which obtains the graph of smaller average degree than the graph obtained by the VG model due to its stricter restrictions and achieved satisfactory results in EEG signal classification. Zhou et al.~\cite{Zhoutt2012lpvg} and Gao et al.~\cite{gaozk2013lpvg} proposed Limited Penetrable Visibility Graph (LPVG), suggesting that the invisible points in VG could also be connected to better process noisy signals, which achieved good performance in two-phase flow pattern identification. Then, Limited Penetrable HVG (LPHVG), as a straightforward extension of HVG, was successfully applied to process EEG signal~\cite{wang2016functional} and electromechanical signal~\cite{wang2016complex}. Besides, the models considering the weights of nodes, representing the importance of points and the direction of edges, and denoting the time irreversibility, have also been proposed~\cite{supriya2016weighted}. The emergence of these VG variants provide different mapping methods for analyzing different types of time series. Such methods have been widely used to solve challenging problems in different research fields, such as finance\cite{yang2009visibility,qian2010universal,wang2012visibility}, physiology \cite{ahmadlou2010new, shao2010network}, meteorology and oceanography \cite{charakopoulos2018dynamics}, geography \cite{elsner2009visibility}, etc.   


In this paper, we try to propose a more flexible mapping method, namely Circular Limited Penetrable Visibility Graph (CLPVG), by introducing nonlinearity into LPVG model. To validate the effectiveness of the model, we first compare the degree distribution of the obtained graphs by CLPVG and LPVG, from two typical time series (periodic, chaotic signals) for qualitative analysis. And then, we utilize these graphs to classify several artificial and real-world time series for quantitative comparison. In the classification experiments, we further expand the structural feature space of the obtained graphs by Subgraph Networks (SGN)~\cite{xuan2019subgraph}. After merging the features extracted from the mapped graphs and their SGNs, we further use Principal Component Analysis (PCA)~\cite{wold1987principal} to reduce dimension and the Random Forest (RF) ~\cite{breiman2001random} as the classifier to realize the time series classification.


The main contributions of this paper are as follows:
\begin{itemize}
\item We propose a new visibility graph model, namely Circular Limited Penetrable Visibility Graph (CLPVG), by introducing nonlinearity into LPVG, to represent time series with a more flexible manner. The experiments on both artificial and real-world time series datasets validate that CLPVG outperforms LPVG in most cases.

\item We utilize Subgraph Network (SGN) for the first time to expand the structural feature space of visibility graphs obtained by LPVG and CLPVG. We find that SGN can indeed enhance both LPVG and CLPVG, in terms of improving their classification accuracy on time series. 

\item To the best of our knowledge, it is also the first time that the visibility graph models, i.e., LPVG and CLPVG, are applied to signal modulation recognition, which achieve reasonable performance and have great potential for further improvement.
\end{itemize}

The rest of this study is organized as follows. In Sec.~\ref{mapping_model}, we introduce our CLPVG, as well as the VG and LPVG models as the basis. In Sec.~\ref{SGN}, we explain how to construct Subgraph Networks (SGN), so as to expand the structural feature space of the obtained visibility graphs. Then, we give the experiments in Sec.~\ref{Exp} and conclude the paper in Sec.~\ref{Con}.

\section{\label{mapping_model}Visibility Graph Model for Time Series}
In this part, we will first introduce the three visibility graph models, including VG, LPVG, and CLPVG. And then, we compare LPVG and CLPVG on artificial time-series datasets.

\subsection{\label{VG}Visibility Graph}
\begin{figure}[!h]
\includegraphics[width=0.45\textwidth]{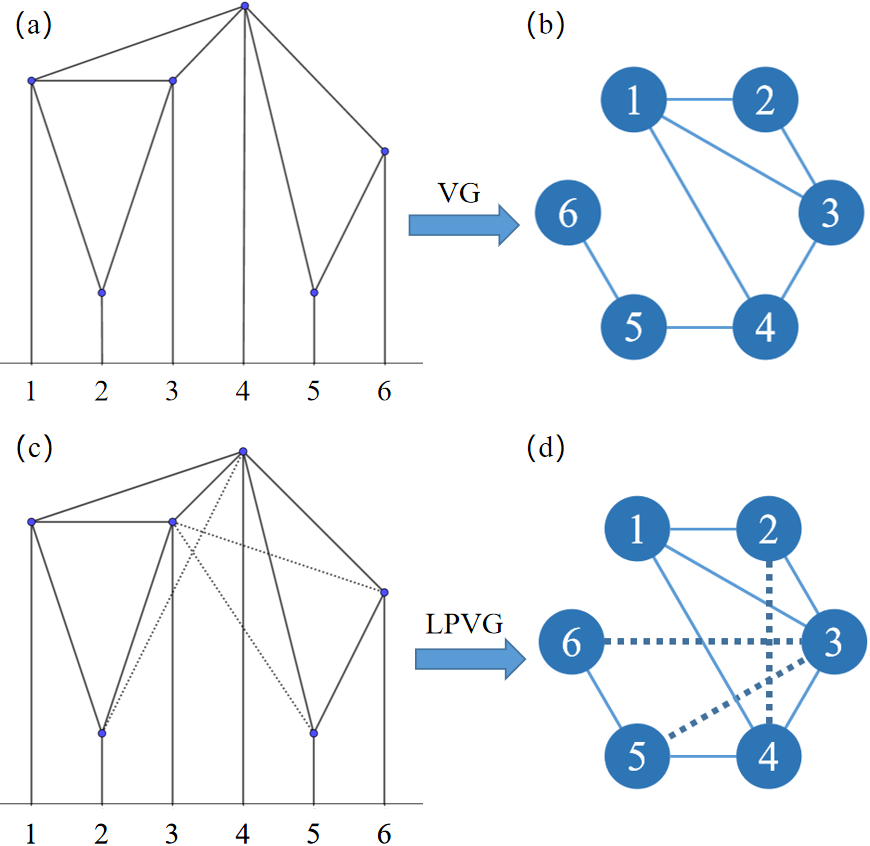}
\caption{\label{fig:VG+LPVG} Schematic diagrams of VG and LPVG ($M$=1).}
\end{figure}

The purpose of the Visibility Graph (VG) model is to build a network from a time series. Then the most important thing is to determine what are the vertices and how they connect. Here, we take a univariate time series $\left\{x_{i}\right\}_{i=1}^{n}$ with $x_{t}=x\left(t_{i}\right)$ for example. Every sampling time point is considered as a vertex, and two vertices $\left(t_{a}, x_{a}\right)$ and $\left(t_{b}, x_{b}\right)$ are connected, if for each sampling point $\left(t_{c}, x_{c}\right)$ with $t_{a}<t_{c}<t_{b}$, it is satisfied that 
\begin{equation}
x_{c}<x_{a}+\left(x_{a}-x_{b}\right)\left(t_{a}-t_{c}\right) /\left(t_{b}-t_{a}\right).
\label{eq:VG}
\end{equation}
The process to establish VG is shown in FIG.~\ref{fig:VG+LPVG} (a)-(b).

\subsection{\label{LPVG}Limited Penetrable Visibility Graph}
Based on VG, Limited Penetrable Visibility Graph (LPVG) is proposed to enhance the anti-noise performance of the visibility graph model. LPVG gives the visibility line for the ability to penetrate, which means that the edges that cannot be connected in VG due to noise may be linked in LPVG. Compared with VG, LPVG defines a limited penetrable visibility distance $M$ additionally. Given any two sampling points $\left(t_{a}, x_{a}\right)$ and $\left(t_{b}, x_{b}\right)$. The visibility criterion in LPVG can be described that if and only if there is no more than $M$ sampling points $\left(t_{c}, x_{c}\right)$ with $t_{a}<t_{c}<t_{b}$ satisfying the criterion:
\begin{equation}
x_{c}>x_{a}+\left(x_{a}-x_{b}\right)\left(t_{a}-t_{c}\right) /\left(t_{b}-t_{a}\right),
\label{eq:LPVG}
\end{equation}
while all the other sampling points satisfy Eq.~(\ref{eq:VG}), then, the mutual connection can be established between them. Based on the LPVG model, the visibility lines represented by the dotted lines shown in FIG.~\ref{fig:VG+LPVG} (c) are newly added, and the constructed graph has three more edges, i.e., (2,4), (3,5) and (3,6), as shown in FIG.~\ref{fig:VG+LPVG} (d).

\subsection{\label{CLPVG}Circular Limited Penetrable Visibility Graph}
In mathematics, a set of circles that meet certain conditions called the circle system, and the equations describing the circle system are the circle system equations.

\begin{figure}[!h]
\includegraphics[width=0.45\textwidth]{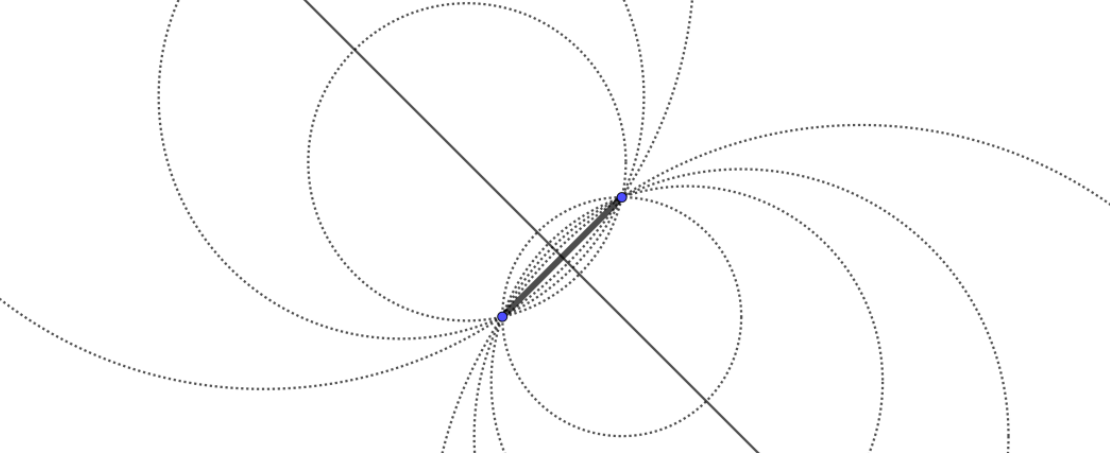}
\caption{\label{fig:circle} Schematic diagram of circle system.}
\end{figure}

As shown in FIG.~\ref{fig:circle}, given any two data points in the time series, we consider selecting the circle set that passes through these two points as the circle system, and the circle system can be described by
\begin{eqnarray}
f(t, x)&=&\left(t-t_{a}\right)\left(t-t_{b}\right)+\left(x-x_{a}\right)\left(x-x_{b}\right) \nonumber\\
&&+\alpha\left[\left(t-t_{a}\right)\left(x_{b}-x_{a}\right)-\left(x-x_{a}\right)\left(t_{b}-t_{a}\right)\right]=0
\label{eq:CLPVG}.
\end{eqnarray}
Based on the idea of the limited penetrating from LPVG which also makes graph construction flexible, we combine circle system and LPVG to propose the Circular Limited Penetrable Visibility Graph (CLPVG). Compared to LPVG, the main difference is that the visibility line is replaced by the visibility arc in CLPVG. Note that here the arc less than 180° is selected to ensure the visibility line meaningful. 

\begin{figure}[!h]
\includegraphics[width=0.45\textwidth]{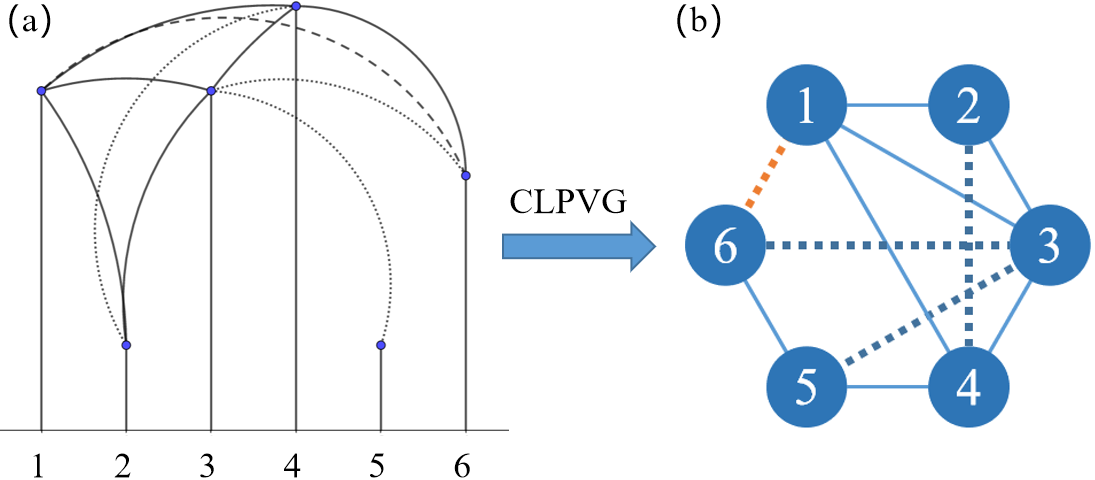}
\caption{\label{fig:CLPVG} Schematic diagram of CLPVG ($M$=1).}
\end{figure}

In particular, given two data points $\left(t_{a}, x_{a}\right)$, $\left(t_{b}, x_{b}\right)$ and the data point $\left(t_{c}, x_{c}\right)$ with $t_{a}<t_{c}<t_{b}$. In CLPVG, denote the $(t_{c}, x_{cir})$ as the point on CLPVG’s visibility arc at $t_{c}$ which satisfies Eq.~(\ref{eq:CLPVG}), and set a limited penetrable visibility distance $M$. Then, take every sampling time point as a vertex first, the same as LPVG, and the edges are created by following the CLPVG rule: if and only if 
no more than $M$ sampling points $\left(t_{c}, x_{c}\right)$ with $t_{a}<t_{c}<t_{b}$ satisfy the criterion $x_{c}>x_{cir}$, while all the other sampling points satisfy $x_{c}<x_{cir}$, the mutual connection is established between $\left(t_{a}, x_{a}\right)$ and $\left(t_{b}, x_{b}\right)$. FIG.~\ref{fig:CLPVG} gives the process to construct CLPVG. Compared with LPVG, there is one more edge, i.e., (1,6), in CLPVG.

\subsection{\label{Model_Assessment}Model Assessment on Artificial Time Series}
To prove the effectiveness and anti-noise performance of our CLPVG model, networks of typical periodic and chaotic time series, as well as their noisy versions, are constructed by LPVG and CLPVG, respectively. The limited penetrated distance $M$ is set to 2 for both models. Then, as usual, we make a simple comparison between the degree distributions of visibility graphs obtained by CLPVG and LPVG, to verify whether the proposed CLPVG can keep the unique information of different types of time series. The experiments show that both models can maintain some characteristics of these time series. And our CLPVG shows higher flexibility of catching features and better ability of anti-noise.

\begin{figure}
\includegraphics[width=0.48\textwidth]{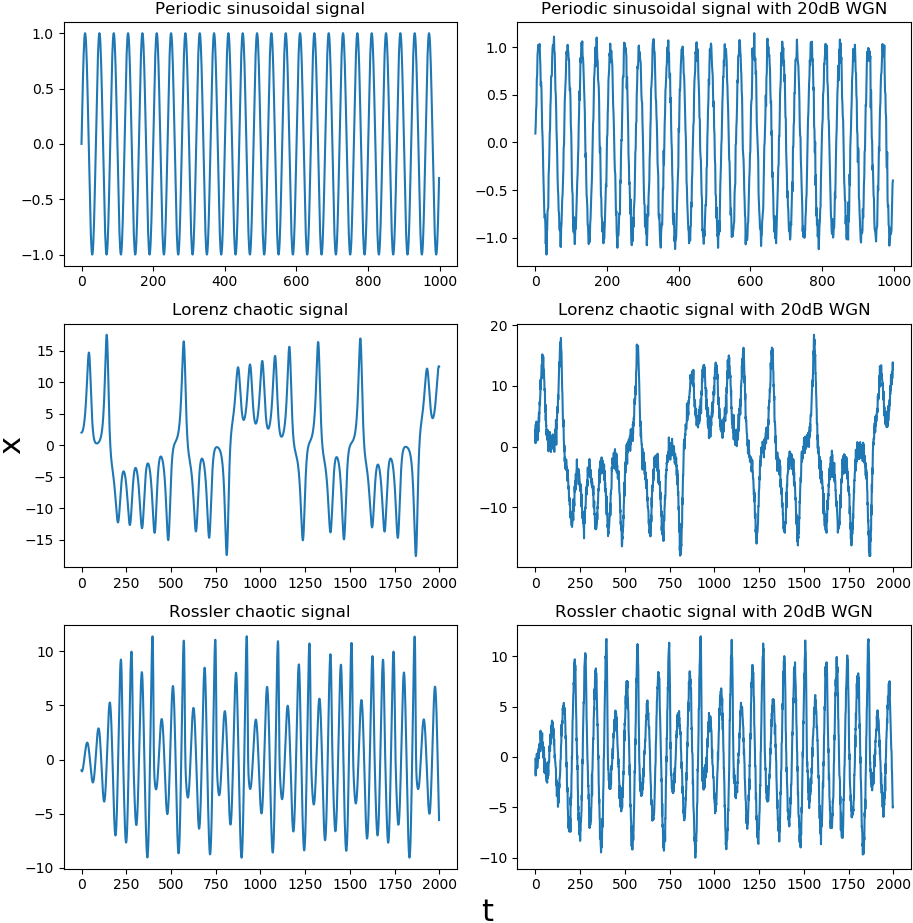}
\caption{\label{fig:PS+CS} The examples of Sinusoidal signal, Lorenz chaotic signal and Rossler chaotic signal, with and without WGN.}
\end{figure}

\begin{figure*}
\includegraphics[width=\textwidth]{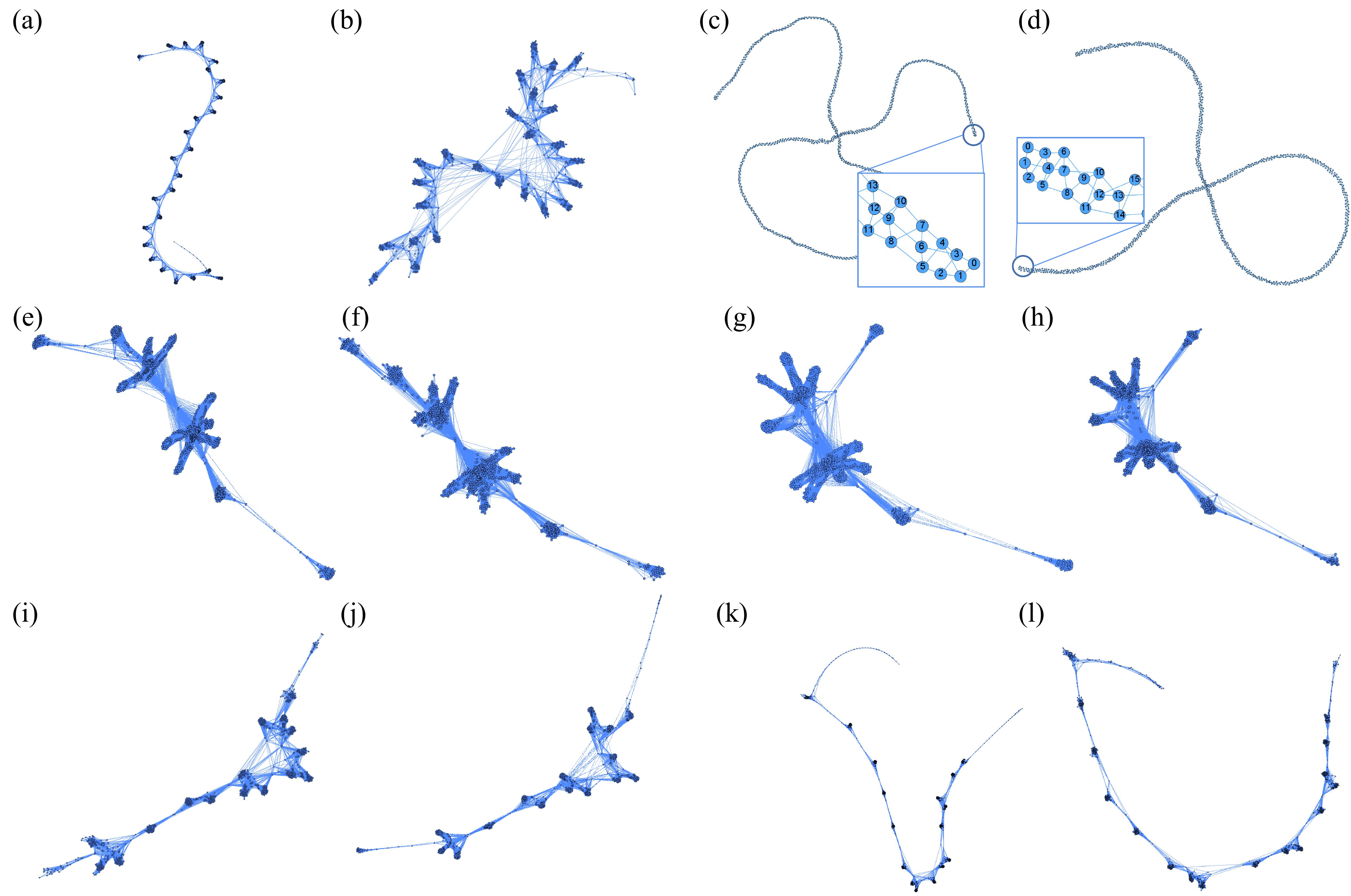}
\caption{\label{fig:net} Graphs constructed by LPVG ($M=2$) and CLPVG ($M=2$), (a)(c): LPVG and CLPVG ($\alpha=1$) of original sinusoidal signal; (b)(d): LPVG and CLPVG ($\alpha=1$) of sinusoidal signal with 20dB WGN; (e)(g): LPVG and CLPVG ($\alpha=10$) of orignal Lorenz signal; (f)(h): LPVG and CLPVG ($\alpha=10$) of Lorenz signal with 20dB; (i)(k): LPVG and CLPVG ($\alpha=10$) of orignal Rossler signal; (j)(l): LPVG and CLPVG ($\alpha=10$) of Rossler signal with 20dB.}
\end{figure*}

In particular, we generate the following three sets of time series: one is periodic and the other two are chaotic.
\begin{itemize}
    \item \textbf{Sinusoidal signal}:
    \begin{equation}
x=\sin (5 \pi t),
\label{eq:PS}
\end{equation}
   \item \textbf{Lorenz chaotic signal}:
\begin{equation}
\left\{\begin{array}{l}
\frac{\mathrm{d} x}{\mathrm{~d} t}=-10(x-y) \\
\frac{\mathrm{d} y}{\mathrm{~d} t}=-y+28 x-x z \\
\frac{\mathrm{d} z}{\mathrm{~d} t}=x y-\frac{8}{3} z
\label{eq:LCS}
\end{array}\right.
\end{equation}
\item \textbf{Rossler chaotic signal}:
\begin{equation}
\left\{\begin{array}{l}
\frac{\mathrm{d} x}{\mathrm{~d} t}=-y-z \\
\frac{\mathrm{d} y}{\mathrm{~d} t}=x+0.2 y \\
\frac{\mathrm{d} z}{\mathrm{~d} t}=0.2+z(x-5.7)
\end{array}\right.
\label{eq:RCS}
\end{equation}
\end{itemize}
For Sinusoidal signal and Lorenz chaotic signal, we sample 1000 points of variable $x$ at 0.01 sampling interval. For Rossler chaotic signal, we sample 1000 points of variable $x$ at 0.1 sampling interval. The initial value of Sinusoidal signal is 0, and those of Lorenz chaotic signal and Rossler chaotic signal are (2, 2, 20) and (-1, 0, 1), respectively. Meanwhile, for each signal, We also add 15dB, 20dB, 30dB and 40dB white gaussian noise (WGN), to see whether our CLPVG is robust to such noises. The examples of Sinusoidal signal, Lorenz chaotic signal, and Rossler chaotic signal, with and without WGN, are shown in FIG.~\ref{fig:PS+CS}. Their corresponding visibility graphs generated by LPVG (the six graphs in the left) and CLPVG (the six graphs in the right) are shown in FIG.~\ref{fig:net}.

\begin{figure*}
\includegraphics[width=\textwidth]{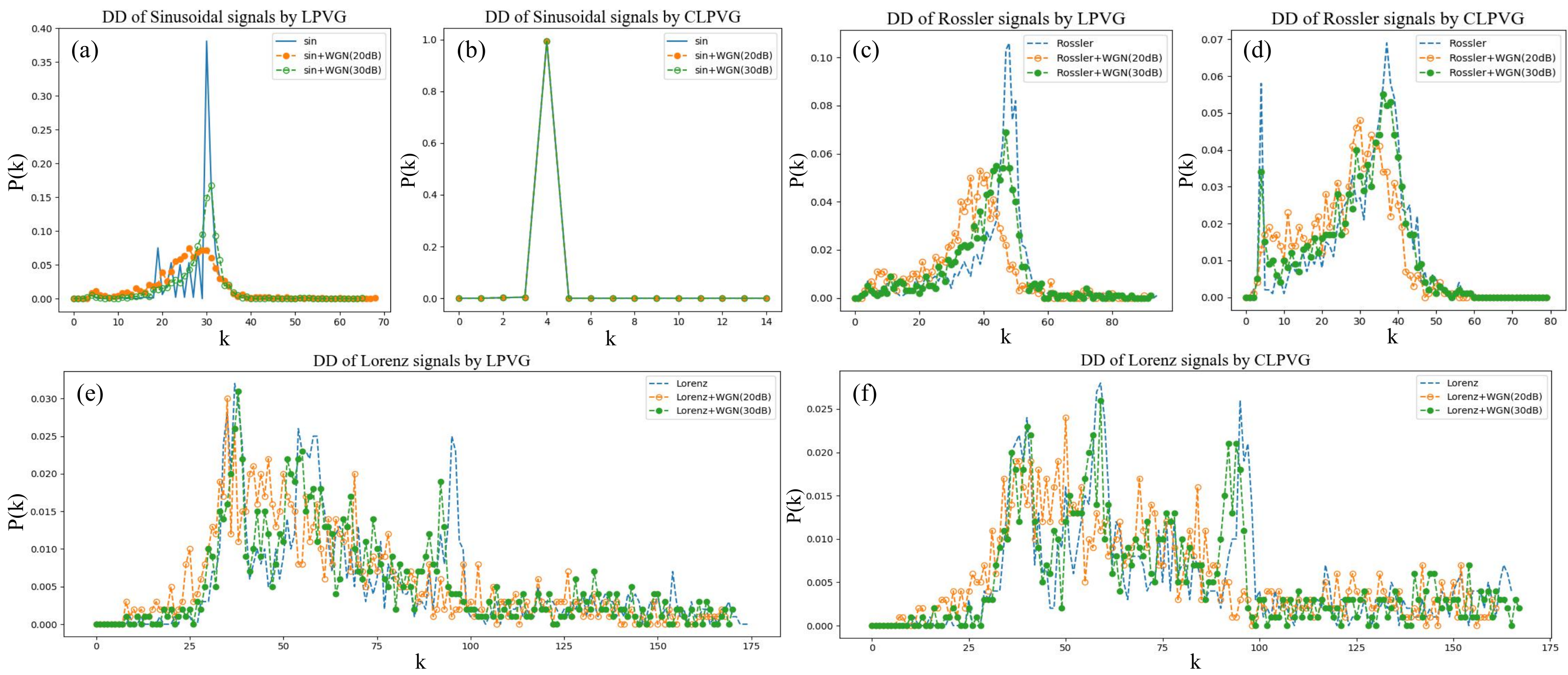}
\caption{\label{fig:PS+CS_DD} The degree distributions (DD) of Sinusoidal signals, Lorenz chaotic signals, and Rossler chaotic signals, with and without WGN.}
\end{figure*}

We present the degree distributions of LPVG and CLPVG, with and without WGN, for three kinds of signals in FIG.~\ref{fig:PS+CS_DD}. Since the Sinusoidal signal is a single-period time series, the degree distributions of LPVG and CLPVG both exhibit regular spikes, as shown in FIG.~\ref{fig:PS+CS_DD} (a)-(b), which proves that these two methods have the ability to distinguish periodic time series. Besides, the tiny change in the position of the main peak in the degree distribution before and after adding WGN means that both LPVG and CLPVG have a certain degree of anti-noise ability. By comparison, there is even no change in the degree distribution of CLPVG before and after adding WGN, indicating its high robustness. 

\begin{table*}[]
\caption{\label{tab:cc}The clustering coefficient of the LPVG and CLPVG on chaotic signals.}
\begin{ruledtabular}
\begin{tabular}{cccccccc}
                       & signal  & orignal & 15dB   & 20dB   & 30dB   & 40dB   & Volatility\footnote{Volatility is the range of ${\left | { CC_{original}- CC_{15dB,20dB,30dB,40dB}} \right|}/CC_{original}$.}    \\ \hline
\multirow{2}{*}{LPVG}  
                       & Rossler & 0.3670  & 0.2940 & 0.3097 & 0.3430 & 0.3610 & 1.63\%-19.89\% \\
                       & Lorenz  & 0.3194  & 0.3252 & 0.2794 & 0.2906 & 0.3125 & 2.16\%-12.52\% \\
\multirow{2}{*}{CLPVG} 
                       & Rossler & 0.2624  & 0.2682 & 0.2672 & 0.2689 & 0.2617 & 0.03\%-2.48\%  \\
                       & Lorenz  & 0.3396  & 0.3033 & 0.3101 & 0.3286 & 0.3348 & 1.41\%-10.68\%
\end{tabular}
\end{ruledtabular}
\end{table*}

Since chaotic attractors have unstable multi-period orbits, the networks built from chaotic signals should have irregular multi-kurtosis distributions. Indeed, for chaotic time series, the degree distributions of LPVG and CLPVG are both irregular and multimodal, as shown in FIG.~\ref{fig:PS&CS_DD} (c)-(f), which means that there is uncertainty in the similarity of node connectivity. Besides, the position of the main peak and the shape of the degree distribution of LPVG and CLPVG from the chaotic signals with WGN don't change much, compared with those without WGN, which means that most noises in chaotic dynamics can be filtered out by LPVG and CLPVG. Since average clustering coefficient is another important property of network structure, we also calculate the average clustering coefficient of LPVG and CLPVG from the chaotic signals with and without WGN, as presented in TABLE~\ref{tab:cc}. It can be clearly found that CLPVG can better preserve the clustering coefficient, i.e., the clustering coefficient of CLPVG change significantly less than that of LPVG, in terms of lower volatility, under various levels of WGN. Again, we can say that CLPVG performs better than LPVG on resisting noises, since the degree distribution peak value, the average clustering coefficient, and the main peak position of CLPVG, before and after adding WGN, are all closer than those of LPVG. 

At last, for these artificial signals, we make a simple test in classifying 300 periodic and 600 chaotic signals with length of 100 by using a graph embedding method Graph2vec~\cite{narayanan2017graph2vec} combined with a machine learning method Random Forest (RF). And the dataset is generated by randomly assigning initial values, the chaotic signals is composed of 300 Lorenz and 300 Rossler signals, and all three types of signals contain the original signals, 20dB and 30dB signals with the same number of samples. The results are obtained by the 10-fold cross-validation, as shown in TABLE~\ref{tab:sin+chaos}. It could still be seen that CLPVG performs better in distinguishing periodic and chaotic signals.

\begin{table}[!t]
\caption{\label{tab:sin+chaos}The accuracy of LPVG ($M=2$) and CLPVG ($M=2$, $\alpha=10$) in classifying periodic and chaotic signals.}
\begin{ruledtabular}
\begin{tabular}{cccc}
                  & Sin vs Chaos     & Sin vs Rossler vs Lorenz & Rossler vs Lorenz \\ \hline
LPVG              & 74.67\%          & 42.89\%                  & 49.17\%           \\
CLPVG             & \textbf{77.33\%} & \textbf{44.44\%}         & \textbf{65.50\%} 
\end{tabular}
\end{ruledtabular}
\end{table}

\section{\label{SGN}Structural Feature Space Expansion}
\begin{figure}[!h]
\includegraphics[width=0.45\textwidth]{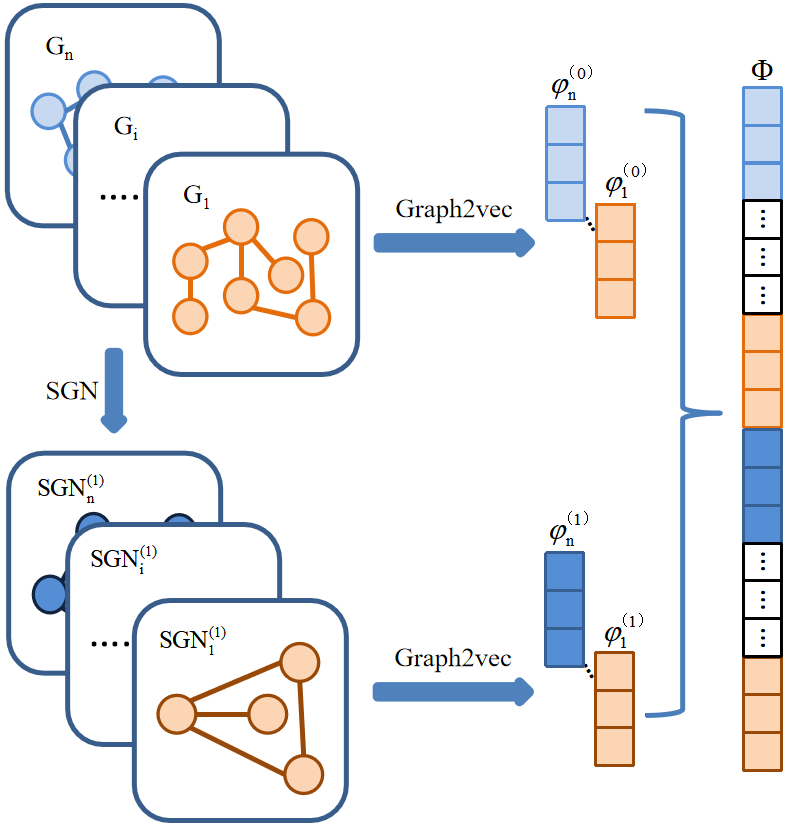}
\caption{\label{fig:fusion_model} The framework of structural feature space expansion.}
\end{figure}

To capture the hidden information of the visibility graphs generated by LPVG and CLPVG, we adopt SGN to expand structural feature space, and further use Graph2vec to automatically extract the structural features, as shown in FIG.~\ref{fig:fusion_model}.  

\subsection{\label{Feature_expanding}Construction of SGN}

\begin{figure}
\includegraphics[width=0.45\textwidth]{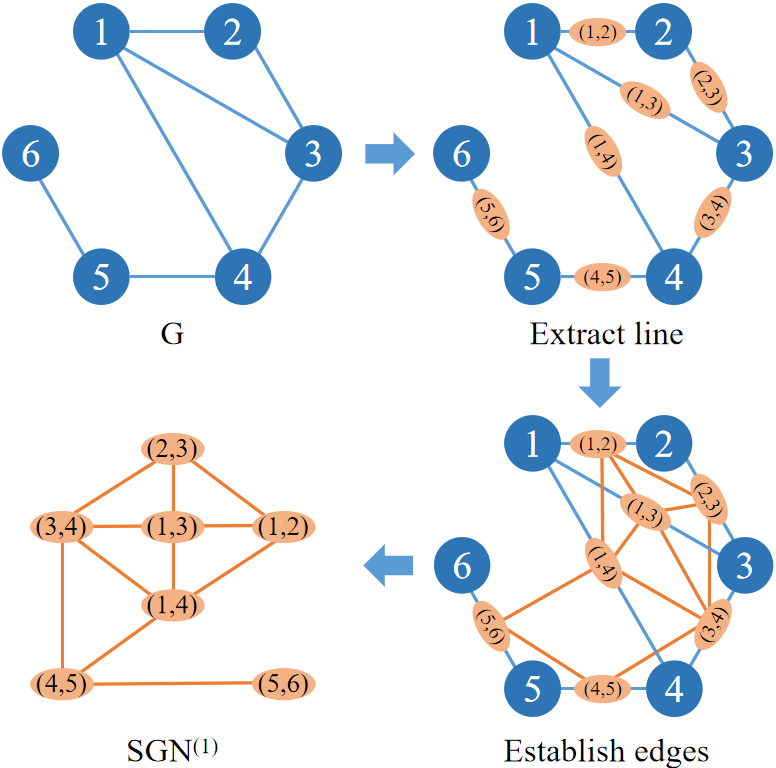}
\caption{\label{fig:SGN} The process of constructing SGN$^{(1)}$ from a given graph $G$.}
\end{figure}

Given an undirected graph $G=\left(V, E\right)$, where $V$ represents the node set and $E$ represents the edge set, we map each edge in $G$ as a node in SGN, and two nodes in SGN are connected if the corresponding edges in $G$ share the same terminal node, so as to form the 1st-order SGN, denoted by SGN$^{(1)}$. The process of constructing 1st-order SGN is shown in FIG.~\ref{fig:SGN}. In general, SGNs of higher orders can be established by iteratively performing the above process, which can provide more structural information for classification. However, the generation of higher-order SGNs is quite time consuming, but contributes less to improve the classification accuracy. Therefore, we only use SGN$^{(1)}$ in this paper.

\begin{figure*}[!t]
\includegraphics[width=0.9\textwidth]{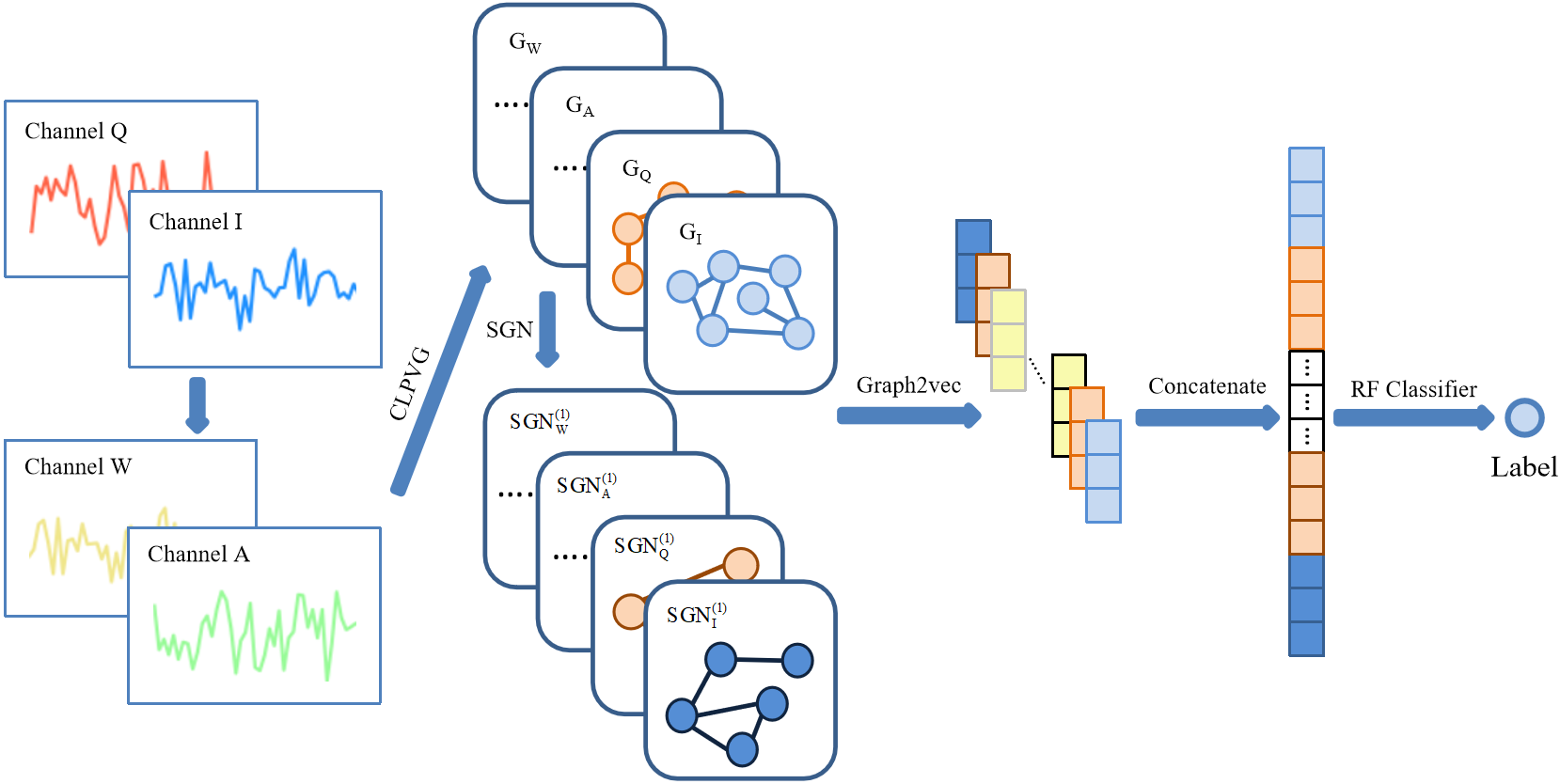}
\caption{\label{fig:exp_frame} The overall framework for radio signal modulation recognition by CLPVG.}
\end{figure*}

\subsection{\label{Feature_extraction}Feature Extraction with Graph2vec}
Here, we adopt Graph2vec~\cite{xuan2019subgraph,narayanan2017graph2vec} to automatically extract structural features of both the original visibility graphs and the corresponding SGN$^{(1)}$. Graph2vec is the first unsupervised embedding method for the entire network, facilitating the subsequent application of machine learning algorithms on graph data. In particular, Graph2vec uses a model similar to Doc2vec~\cite{le2014distributed} to establish a relationship between the network and the rooted subgraph. It first extracts the rooted subgraph and provides the corresponding label into the vocabulary, and then trains the Skip-Gram model to obtain the representation of the entire network. Numerous experiments validate that Graph2vec performs well in many graph classification tasks~\cite{xuan2019subgraph}.

\subsection{\label{Feature_Fusion}Feature Fusion}
Usually, a signal may consist of multiple channels like RGB channels in image. For example, a modulated signal of I/Q digital modulation has channel $I$ and channel $Q$. Besides, other characteristics in time domain, such as amplitude $A$ and phase $W$, can be established from the original channels. Generally, a signal $S$ can be represented by $S=\left\{S_{i}\right\}_{i=1}^{n}$ and the $i$-th channel is represented by $S_{i}$. The set of graphs mapped from all channels is denoted by $G=\left\{G_{i}\right\}_{i=1}^{n}$, in which $G_{i}$ represents the graph mapped from the $i$-th channel of the signal. The set of the 1st-order SGNs mapped from all channels is denoted by $\mathrm{SGN}^{(1)}=\left\{\mathrm{SGN}_{i}^{(1)}\right\}_{i=1}^{n}$, in which SGN$_i^{(1)}$ represents the 1st-order SGN extracted from $G_{i}$. For each $G_{i}$ and SGN$_i^{(1)}$, we can get their feature vectors by Graph2vec as follows:
\begin{equation}
\varphi_{i}^{(0)}=\text { Graph2vec }\left(G_{i}\right),
\label{eq:graph2vec_G}
\end{equation}
\begin{equation}
\varphi_{i}^{(1)}=\operatorname{Graph} 2 \operatorname{vec}\left(\mathrm{SGN}_{i}^{(1)}\right),
\label{eq:graph2vec_SGN1}
\end{equation}
with $\varphi_{i}^{(0)} \in R^{K}$ and $\varphi_{i}^{(1)} \in R^{K}$. Then, all the extracted feature vectors are fused together to obtain a single vector, denoted by $\Phi \in R^{2 n K}$:
\begin{equation}
\Phi=\varphi_{1}^{(0)} \| \varphi_{2}^{(0)} \ldots \| \varphi_{n}^{(0)} \|\varphi_{1}^{(1)} \| \varphi_{2}^{(1)} \ldots \| \varphi_{n}^{(1)},
\end{equation}
where $\|$ represents a fusion operation in the horizontal direction. The simple fusion method results in higher dimensionality of the feature vector, therefore Principal Components Analysis (PCA) is then adopted to reduce the dimensionality of the unified feature vector, to retain the features that contribute the most to the variance. After PCA processing, we get a new $\theta$-dimensional feature vector:
\begin{equation}
\Phi_{\theta}=\text{PCA}(\Phi),
\end{equation}
where $\Phi_{\theta} \in R^{\theta}$. This feature vector then is used as the input of RF classifier.

\section{\label{Exp}Experiments on Real-World Datasets}
We apply LPVG and CLPVG on radio signal modulation classification and EEG signal epilepsy detection to testify their effectiveness. The overall framework for radio signal modulation classification is shown in FIG.~\ref{fig:exp_frame}, and the training set and test set are separated from the original dataset by 4:1. For EEG, we just need to fuse the features of visibility graph and its corresponding SGN$^{(1)}$, and the results are obtained by 10-fold cross-validation. At first, we convert the time series into the visibility graphs by LPVG and CLPVG, respectively, with the limited penetrated distance $M$ set to 1. Then, we apply SGN model to expand the structural feature space for the visibility graph of each channel. Finally, we use feature fusion and PCA to get a unified feature vector, and then use RF classifier to realize the classification.

\subsection{\label{Dataset}Datasets}
In this paper, we use two real-world datasets which are described in the following. 
\begin{itemize}
    \item \textbf{RADIOML 2016.10A}~\cite{o2016convolutional}: It is a synthetic dataset generated by GNU Radio, which was first released at the 6th GNU Annual Radio Conference. This signal dataset contains 11 modulation types (BPSK, QPSK, 8PSK, 16QAM, 64QAM, BFSK, CPFSK, and PAM4 for 8 digital modulations, and WB-FM, AM-SSB, and AM-DSB for 3 analog modulations). Each modulation signal contains 20 different signal-to-noise ratios (SNR). Each SNR contains 1000 samples. Each sample has two quadrature signals $I/Q$, and each signal contains 128 sampling points. 
    \item \textbf{Epilepsy EEG}~\cite{supriya2016weighted}: It is published by the Department of Epilepsy at the University of Bonn. All EEG recordings use the same 128-channel amplifier system and use the average common reference value for recording. The recorded data is digitized at 173.61 samples per second with a 12-bit resolution, and the bandpass filter is set to 0.53 Hz to 85 Hz. The complete EEG database contains five categories, which are represented as A, B, C, D and E. The first two are the surface records of healthy volunteers with eyes opened and closed, respectively. C is the intracranial records of patients' brain in seizure free interval during the hippocampal formation of the opposite hemisphere. D is the intracranial record of epileptic patient's epileptogenic zone in seizure free interval. E is the record during seizure activity. Each category has 100 signals, and each signal has a length of 4096. A 10-20 electrode placement system is used to record EEG signals.
\end{itemize}
\subsection{\label{Data_prepocessing}Data Preprocessing}
We first use peak detection~\cite{palshikar2009simple} to compress time series, with the aim of reducing time complexity and noise. In particular, given a time series $S=\{x_1, x_2, \cdots , x_n\}$ a hyper-parameter $w$ is denoted as time-window size. First, we extend the original time series $S$ to $S^1=\{0, 0, \cdots , 0, x_1, x_2, \cdots , x_n, 0, 0, \cdots , 0\}$ by adding $w$ zero elements at the beginning and the end of $S$. Then, we intercept $x_k$’s left segment $\{x_{k-w}, \cdots , x_{k-1}, x_{k}\}$ and right segment $\{x_{k}, x_{k+1}, \cdots , x_{k+w}\}$ with length $(w+1)$, and mark their maximum as $x_{\text {left-max}}$ and $x_{\text {right-max}}$, respectively. Finally, $x_k (k\in [1, n])$ will be retained if $\left(x_{\text {left-max}}+x_{\text {right-max} }\right) / 2 \leq x_{k}$ or else removed from the sequence.

For RADIOML 2016.10A, we use amplitude $A$ and phase $W$, to expand the original signals, where $A=\sqrt{I^{2}+Q^{2}}$ and $W=\arctan \frac{Q}{I}$, as shown in FIG.~\ref{fig:exp_frame}. Then, a complete dataset with a size of 220,000×4×128 can be obtained for the subsequent use. We further adopt the above peak detection algorithm with the window size of 3 and 4 to further compress the signals. For Epilepsy EEG, it could be too complex to convert each 4096-length time series into a graph. Therefore, we divide each 4096-length signal into 4 segments of equal length 1024, so as to increase the number of samples. According to the reference~\cite{supriya2016weighted}, there is little difference in accuracy when considering segmented and non-segmented EEG signals. Then, we use the peak detection algorithm with the window size of 3 to further compress the signals.

\begin{table*}
\caption{\label{tab:table1}The accuracy of the LPVG ($M=1$) and CLPVG ($M=1$, $\alpha=10$) on data set RML2016.10a without SGN$^{(1)}$.}
\begin{ruledtabular}
\begin{tabular}{ccccccccccccc}
          & \multicolumn{6}{c}{without SGN$^{(1)}$}                                                                                                                                                                            & \multicolumn{6}{c}{with SGN$^{(1)}$}                                                                            \\
          & \multicolumn{2}{c}{$w=3$}           & \multicolumn{2}{c}{$w=4$}           & \multicolumn{2}{c}{$merge(3-4)\footnote{$merge(a-b)$ means merging the extracted feature matrix of window size between $a$ and $b$.}$} & \multicolumn{2}{c}{$w=3$}           & \multicolumn{2}{c}{$w=4$}           & \multicolumn{2}{c}{$merge(3-4)$}    \\ 
SNR                  & LPVG             & CLPVG            & LPVG             & CLPVG            & LPVG                                                                 & CLPVG                                                               & LPVG             & CLPVG            & LPVG             & CLPVG            & LPVG             & CLPVG            \\ \hline
18dB      & 74.14\%          & \textbf{76.18\%} & 78.14\%          & \textbf{80.64\%} & 75.68\%                                                            & \textbf{78.41\%}                                                  & 76.00\%          & \textbf{76.95\%} & \textbf{80.77\%} & 79.35\%          & \textbf{81.68\%} & 80.63\%          \\
16dB      & 73.27\%          & \textbf{74.50\%} & 75.64\%          & \textbf{80.64\%} & 75.18\%                                                            & \textbf{78.55\%}                                                  & 76.64\%          & \textbf{78.18\%} & \textbf{79.50\%} & 78.83\%          & \textbf{80.73\%} & 79.96\%          \\
14dB      & 72.77\%          & \textbf{73.82\%} & 74.82\%          & \textbf{77.95\%} & 73.73\%                                                            & \textbf{76.32\%}                                                  & 76.00\%          & \textbf{76.50\%} & \textbf{78.86\%} & 78.76\%          & \textbf{80.05\%} & 79.44\%          \\
12dB      & 71.23\%          & \textbf{73.73\%} & 76.59\%          & \textbf{78.05\%} & 73.77\%                                                            & \textbf{77.27\%}                                                  & 73.91\%          & \textbf{75.32\%} & \textbf{80.41\%} & 78.03\%          & \textbf{79.36\%} & 79.20\%          \\
10dB      & 73.77\%          & \textbf{75.64\%} & 74.45\%          & \textbf{79.00\%} & 74.68\%                                                            & \textbf{77.36\%}                                                  & 74.95\%          & \textbf{76.41\%} & 79.36\%          & \textbf{79.44\%} & 78.95\%          & \textbf{79.05\%} \\
8dB       & 71.77\%          & \textbf{74.09\%} & 73.91\%          & \textbf{76.18\%} & 73.09\%                                                            & \textbf{75.32\%}                                                  & 73.23\%          & \textbf{74.82\%} & \textbf{77.05\%} & 76.21\%          & \textbf{77.64\%} & 76.87\%          \\
6dB       & 67.09\%          & \textbf{70.55\%} & 69.18\%          & \textbf{73.41\%} & 70.05\%                                                            & \textbf{72.73\%}                                                  & 69.05\%          & \textbf{71.64\%} & \textbf{74.82\%} & 73.44\%          & \textbf{74.23\%} & 72.23\%          \\
4dB       & 63.05\%          & \textbf{67.77\%} & 64.86\%          & \textbf{67.64\%} & 67.05\%                                                            & \textbf{68.91\%}                                                  & 64.45\%          & \textbf{71.00\%} & 66.55\%          & \textbf{67.48\%} & 67.55\%          & \textbf{71.75\%} \\
2dB       & \textbf{51.09\%} & 50.82\%          & 52.95\%          & \textbf{55.09\%} & \textbf{55.73\%}                                                   & 54.91\%                                                           & 52.32\%          & \textbf{62.82\%} & 55.86\%          & \textbf{59.97\%} & 58.05\%          & \textbf{62.57\%} \\
0dB       & 41.27\%          & \textbf{43.91\%} & 44.09\%          & \textbf{48.36\%} & \textbf{49.45\%}                                                   & 47.50\%                                                           & 43.59\%          & \textbf{56.68\%} & 46.55\%          & \textbf{57.15\%} & 46.68\%          & \textbf{58.33\%} \\
-2dB      & 32.45\%          & \textbf{33.91\%} & 38.82\%          & \textbf{38.86\%} & \textbf{41.18\%}                                                   & 37.45\%                                                           & 37.50\%          & \textbf{46.82\%} & 42.32\%          & \textbf{48.45\%} & 40.59\%          & \textbf{48.30\%} \\
-4dB      & \textbf{32.14\%} & 31.77\%          & \textbf{37.36\%} & 35.23\%          & \textbf{37.86\%}                                                   & 35.64\%                                                           & 33.77\%          & \textbf{39.73\%} & \textbf{41.55\%} & 39.63\%          & 39.82\%          & \textbf{40.04\%} \\
-6dB      & 32.59\%          & \textbf{32.73\%} & 34.09\% & \textbf{34.14\%}          & \textbf{36.14\%}                                                   & 35.45\%                                                           & 30.59\%          & \textbf{33.95\%} & \textbf{38.45\%} & 38.12\%          & \textbf{38.45\%} & 35.49\%          \\
-8dB      & \textbf{29.55\%} & 29.32\%          & 27.05\%          & \textbf{30.23\%} & \textbf{31.64\%}                                                   & 30.77\%                                                           & 25.55\%          & \textbf{28.18\%} & 29.05\%          & \textbf{29.41\%} & \textbf{30.86\%} & 29.03\%          \\
-10dB     & 24.82\%          & \textbf{26.09\%} & 20.77\%          & \textbf{23.68\%} & 24.91\%                                                            & \textbf{25.23\%}                                                  & \textbf{21.23\%} & 21.00\%          & \textbf{19.45\%} & 19.27\%          & \textbf{23.27\%} & 22.01\%          \\
-12dB     & 21.86\% & \textbf{21.86\%} & 17.59\%          & \textbf{19.45\%} & \textbf{21.86\%}                                                   & 21.18\%                                                           & \textbf{20.23\%} & 19.45\%          & \textbf{17.59\%} & 16.94\%          & \textbf{19.64\%} & 19.32\%          \\
-14dB     & 20.09\%          & \textbf{20.45\%} & 16.00\%          & \textbf{16.36\%} & \textbf{20.73\%}                                                   & 20.45\%                                                           & 17.95\%          & \textbf{18.82\%} & 15.91\%          & \textbf{16.56\%} & 18.14\%          & \textbf{19.36\%} \\
-16dB     & \textbf{18.50\%} & 18.36\%          & 16.41\%          & \textbf{16.50\%} & \textbf{19.23\%}                                                   & 17.55\%                                                           & 16.68\%          & \textbf{18.86\%} & \textbf{14.86\%} & 14.41\%          & 17.45\%          & \textbf{17.93\%} \\
-18dB     & \textbf{19.55\%} & 19.18\%          & 15.55\%          & \textbf{16.68\%} & \textbf{19.27\%}                                                   & 18.73\%                                                           & \textbf{17.55\%} & 17.36\%          & 15.32\%          & \textbf{15.78\%} & 16.91\%          & \textbf{17.61\%} \\
-20dB     & \textbf{18.91\%} & 18.50\%          & 16.14\%          & \textbf{16.36\%} & \textbf{18.23\%}                                                   & 17.64\%                                                           & \textbf{17.00\%} & 16.95\%          & \textbf{15.36\%} & 15.33\%          & 16.82\%          & \textbf{16.90\%} \\ \hline
$>$ 5dB   & 72.01\%          & \textbf{74.07\%} & 74.68\%          & \textbf{77.98\%} & 73.74\%                                                            & \textbf{76.56\%}                                                  & 74.25\%          & \textbf{75.69\%} & \textbf{78.68\%} & 77.72\%          & 78.95\%          & \textbf{79.20\%} \\
$\ge$ 0dB & 65.95\%          & \textbf{68.10\%} & 68.46\%          & \textbf{71.70\%} & 68.84\%                                                            & \textbf{70.73\%}                                                  & 68.01\%          & \textbf{72.03\%} & 71.97\%          & \textbf{72.86\%} & 72.49\%          & \textbf{74.00\%} \\
$>$ -5dB  & 60.34\%          & \textbf{62.22\%} & 63.40\%          & \textbf{65.92\%} & 63.95\%                                                            & \textbf{65.03\%}                                                  & 62.62\%          & \textbf{67.24\%} & 66.97\%          & \textbf{68.06\%} & 67.11\%          & \textbf{68.62\%} \\
Total     & 45.50\%          & \textbf{46.66\%} & 46.22\%          & \textbf{48.22\%} & 47.97\%                                                            & \textbf{48.37\%}                                                  & 45.91\%          & \textbf{49.07\%} & 48.48\%          & \textbf{49.13\%} & 49.34\%          & \textbf{50.30\%}

\end{tabular}

\end{ruledtabular}
\end{table*}

\subsection{\label{Results}Results and Discussion}
We first compare LPVG and CLPVG on the radio signal dataset, and simply set $\alpha=10$ for all the channels. The experimental results under different SNR are shown in TABLE~\ref{tab:table1}.
To ensure a fair comparison between LPVG and CLPVG, we test their classification accuracy under the same experimental setting. Overall, we can get higher accuracy by using CLPVG, rather than LPVG, and such superiority is more obvious for the signals of higher SNR. This indicates that CLPVG is able to extract more effective structural information from signals of relatively high SNR, by setting an appropriate hyper-parameter $\alpha$. In addition to setting a same hyper-parameter for all channels of signals, we can also set different $\alpha$ for different channels to extract information unique to this channel and then fuse them to get a better classification accuracy. However, the optimization of hyper-parameters may be time consuming, which is left for future expansion. We also find that the time-window size may have an effect on the accuracy, and such effect may vary for different SNR, suggesting that we can choose appropriate time-window size to improve the classification accuracy of signals with specific SNR. 

Moreover, we further expand the structural features of visibility graphs by SGN to improve the classification accuracy. Here, only the 1st-order SGN is applied, and the results are also shown in TABLE~\ref{tab:table1}. We find that SGN can indeed enhance both LPVG and CLPVG, i.e., the classification accuracy is significantly higher when SGN is adopted. By comparison, LPVG benefits more from SGN, indicating that LPVG is less representative effective than CLPVG on capturing signal latent characteristics, while such weakness could be overcome to certain extent by mapping the visibility graphs to higher-order networks, e.g., SGN. Nevertheless, SGN enhanced CLPVG still behaves better than SGN enhanced LPVG, validating again the superiority of CLPVG over LPVG. We also merge the feature vectors obtained by different time windows, and find that the overall accuracy could be further improved for SGN enhanced CLPVG.

We also compare LPVG and CLPVG on the Epilepsy EEG dataset, with the results shown in TABLE~\ref{tab:table3}. Since there are totally five categories in this dataset, here we design two classification tasks: classify E from the rest, and classify all the five categories from each other. For the first easy task, both LPVG and CLPVG behave quite well, achieving the accuracy higher than 97\%, and CLPVG is slightly better than LPVG in this case. For the second difficult task, CLPVG achieves significantly higher accuracy than LPVG, especially when SGN is not adopted to expand the feature space, validating the effectiveness of our CLPVG model. 

\begin{table}
\begin{ruledtabular}
\caption{\label{tab:table3}The accuracy of LPVG ($M=1$) and CLPVG ($M=1$, $\alpha=10$), with and witout SGN, on Epilepsy EEG dataset.}
\begin{tabular}{llccc}
                              & \multicolumn{2}{c}{without SGN$^{(1)}$}             & \multicolumn{2}{c}{with SGN$^{(1)}$} \\
\multicolumn{1}{c}{Task}            & LPVG     & CLPVG            & LPVG     & CLPVG             \\ \hline
\multicolumn{1}{c}{ABCD-E}    & 97.75\%                  & \textbf{97.80\%} & 97.60\%  & \textbf{97.80\%}  \\
\multicolumn{1}{c}{A-B-C-D-E} & 64.82\%                  & \textbf{67.40\%} & 72.62\%  & \textbf{73.42\%} 
\end{tabular}
\end{ruledtabular}
\end{table}

\section{\label{Con}Conclusion}
Visibility graph provides a new way for signal processing. In this paper, we introduce a new approach to construct visibility graphs, namely Circular Limited Penetrated Visibility Graph (CLPVG). To the best of our knowledge, this is the first time to introduce circle system into the construction of visibility graphs, which increases the flexibility of our CLPVG model. A series of experiments on both artificial and real-world time-series datasets validate the better anti-noise ability of our CLPVG, compared with LPVG, leading to its higher performance on time-series classification. 

This study may trigger a burst of study on visibility graph by introducing more various nonlinear mapping mechanisms beyond circle system. It should be noted that, here we only focus on proposing more flexible visibility graph models, and thus just compare our CLPVG with classic LPVG to validate its effectiveness. Such models establish a bridge between time series and graphs, help to better understand the structure of time series by visualizing them with graphs, and may inspire researchers to propose more effective graph algorithms for signal processing. In the future, such visibility graph models could be integrated into deep learning frameworks~\cite{chen2020signet} to automatically generate graphs for time series, which is expected to significantly improve their feature extraction ability.

\begin{acknowledgments}
This work was partially supported by the National Natural Science Foundation of China under Grant No. 61973273, and by the Zhejiang Provincial Natural Science Foundation of China under Grant No. LR19F030001.
\end{acknowledgments}

\section*{data availablity} 
The data that supports the findings of this study are available within the article and their corresponding references.

\section*{reference} 
\bibliography{zhou_reference}
\end{document}